\documentclass[]{saisart}

\usepackage{listings}
\usepackage[utf8]{inputenc}
\usepackage[T1]{fontenc}
\usepackage{graphicx}
\usepackage{url}
\usepackage{hyperref}
\usepackage{xcolor}
\usepackage{textcomp,gensymb}

\urldef{\footurlAirSwimmers}\url{https://en.wikipedia.org/wiki/Air_Swimmer}
\urldef{\footurlNanoBlimp}\url{https://ohgizmo.com/nanoblimp-is-allegedly-worlds-smallest-rc-blimp}
\urldef{\footurlAirRay}\url{https://www.youtube.com/watch?v=c3-wIICjAhE}
\urldef{\footurlAirJelly}\url{https://www.youtube.com/watch?v=divLsTtA5vk}
\urldef{\footurlAirPenguin}\url{https://www.youtube.com/watch?v=jPGgl5VH5go}
\urldef{\footurlSmartInversion}\url{https://www.youtube.com/watch?v=BDiTicX8HRU}
\urldef{\footurlFreeMotionHandling}\url{https://www.youtube.com/watch?v=dv6zQr1_C9Q}
\urldef{\footurlEmotionSpheres}\url{https://www.youtube.com/watch?v=5iqP1oPZ3Qw}
\urldef{\footurlRPI}\url{https://www.raspberrypi.com/products/raspberry-pi-zero-2-w}
\urldef{\footurlESP}\url{https://www.espressif.com/en/products/modules/esp32}
\urldef{\footurlLuna}\url{https://www.waveshare.com/wiki/TF-Luna\_LiDAR\_Range\_Sensor}
\urldef{\footurlThermal}\url{https://www.melexis.com/en/product/mlx90640/far-infrared-thermal-sensor-array}
\urldef{\footurlHelium}\url{https://specialtygases.messergroup.com/balloon-helium}
\urldef{\footurlWriter}\url{https://perchance.org/znevive9bn}
\urldef{\footurlFiber}\url{https://listverse.com/2022/03/05/10-criminals-who-were-caught-thanks-to-trace-fiber-evidence/}
\urldef{\footurlParticles}\url{https://www.nbcnews.com/news/us-news/gary-ridgway-green-river-serial-killer-washington-rcna67794}
\urldef{\footurlUtah}\url{https://www.blimpinfo.com/uncategorized/plans-for-ogden-police-blimp-canceled/}
\urldef{\footurlAngel}\url{https://www.youtube.com/watch?v=4vbQ2tMOjgI/}
\urldef{\footurlBlood}\url{https://www.forensicsciencesimplified.org/blood/principles.html}
\urldef{\footurlTigerVNC}\url{https://tigervnc.org/}
\urldef{\footurlOpenCV}\url{https://opencv.org/}
\urldef{\footurlYOLO}\url{https://docs.ultralytics.com/}
\urldef{\footurlChatGPT}\url{https://chatgpt.com}
\urldef{\footurlYouTube}\url{https://youtu.be/5ep9UeChn68}
\urldef{\footurlGitHub}\url{https://github.com/martincooney/Crime_Blimp}
\urldef{\footurlArduino}\url{https://www.arduino.cc}
\urldef{\footurlLiPoShim}\url{https://shop.pimoroni.com/products/lipo-shim}

\urldef{\footurlContamination}\url{https://universalclass.com/articles/law/setting-crime-scene-perimeters.htm}
\urldef{\footurlDefinition}\url{https://www.airships.net/dirigible}

\urldef{\footurlESPWeight}\url{https://blog.arducam.com/esp32-machine-vision-learning-guide/}
\urldef{\footurlESPCurrent}\url{https://docs.sunfounder.com/projects/galaxy-rvr/en/latest/hardware/cpn_esp_32_cam.html}
\urldef{\footurlRPIWeight}\url{https://picockpit.com/raspberry-pi/everything-about-raspberry-pi-zero-2-w/}
\urldef{\footurlRPICurrent}\url{https://andrejacobs.org/b24/electronics/raspberry-pi-zero-2-w-temperature-and-power-consumption/}
\urldef{\footurlMLXSpecs}\url{https://www.adafruit.com/product/4407}
\urldef{\footurlLunaSpecs}\url{https://www.waveshare.com/wiki/TF-Luna_LiDAR_Range_Sensor}
\urldef{\footurlLiPoShimSpecs}\url{https://www.pishop.us/product/pimoroni-zero-lipo-shim/}
\urldef{\footurlOpenMV}\url{https://openmv.io}
\urldef{\footurlOpenMVTwo}\url{https://openmv.io/collections/all-products/products/openmv-n6}
\urldef{\footurlLLM}\url{https://pyvideotrans.com/en/_posts/deployllm}

\urldef{\footurlZephyr}\url{https://www.3dflow.net/3df-zephyr-photogrammetry-software/}

\begin{document}

\copyrightyear{2025}
\copyrightclause{Copyright for this paper by its authors.
  Use permitted under Creative Commons License Attribution 4.0
  International (CC BY 4.0).}

\conference{SAIS2025: Swedish AI Society Workshop 2025, 16-17 June 2025, Halmstad, Sweden.}

%%
%% The "title" command: please use "title case capitalisation" 
%% Guidelines can be found here: https://wiki.musicbrainz.org/Style/Language/English
\title{Blimp-based Crime Scene Analysis*}
%\tnotetext[1]{You can use this document as the template for preparing your publication. We recommend using the latest version of the saisart style.}

\author[1]{Martin Cooney}[%
orcid=0000-0002-4998-1685,
email=martin.daniel.cooney@gmail.com,
url=https://martindanielcooney.wordpress.com,
]
\cormark[1]
\address[1]{School of Information Technology, Halmstad University, Kristian IV:s väg 3, Halmstad
301 18 Sweden}
%\address[2]{Physical address, second affiliation}

\author[1]{Fernando Alonso-Fernandez}[%
orcid=0000-0002-1400-346X,
email=fernando.alonso-fernandez@hh.se,
%url=https://www.sais.se/,
]
%\address[3]{Physical address, third affiliation}

%\author[4]{Third Author}[%
%orcid=0000-0000-0000-0000,
%email=email@gmail.com,
%url=https://www.sais.se/,
%\address[4]{Physical address, fourth affiliation}

\cortext[1]{Corresponding author.}
%\fntext[1]{These authors contributed equally.}

\begin{abstract}
Crime is a critical problem---which often takes place behind closed doors, posing additional difficulties for investigators.
To bring hidden truths to light, evidence at indoor crime scenes must be documented before any contamination or degradation occurs.
Here, we address this challenge from the perspective of artificial intelligence (AI), computer vision, and robotics:
Specifically, we explore the use of a blimp as a "floating camera" to drift over and record evidence with minimal disturbance.
Adopting a rapid prototyping approach, we develop a proof-of-concept to investigate capabilities required for manual or semi-autonomous operation.
Consequently, our results demonstrate the feasibility of equipping indoor blimps with various components (such as RGB and thermal cameras, LiDARs, and WiFi, with 20 minutes of battery life). Moreover, we confirm the core premise: that such blimps can be used to observe crime scene evidence while generating little airflow. We conclude by proposing some ideas related to detection (e.g., of bloodstains), mapping, and path planning, with the aim of stimulating further discussion and exploration.
\end{abstract}

\begin{keywords}
  small blimp \sep
  indoor crime scene analysis \sep
  exploratory design \sep
  applied AI
\end{keywords}

\maketitle

%The United Nations state a clear need for leveraging technologies to invest in societal safety from crime, which is a fundamental requirement for social and economic development, in line with sustainable development goal (SDG) 16~\cite{un2023sdg}.
%Although the degree to which crime ravages lives cannot be easily quantified, the same report highlights, for example, that intentional homicides cruelly took approximately half of a million lives in 2021---the highest rate in twenty years.
%Economically, the annual global damage is also assessed to be a massive \$9.4 trillion USD, with homicide at 705.89 billion USD and assaults at 325.27 billion USD~\cite{hoeffler2017costs}.

\section{Introduction}
%\section{INTRODUCTION}
\label{section:intro}

At the crossroads of applied AI, design, computer vision and robotics, this paper explores how a blimp could be used to non-disruptively analyze indoor crime scenes.

Crime remains a global crisis, ruining numerous lives and draining public resources: 
Intentional homicides cruelly took approximately half of a million lives in 2021---the highest rate in twenty years~\cite{un2023sustainable}.
The global cost of violence has also been estimated at a staggering \$9.4 trillion USD, with homicide at \$705.89 billion and assaults at \$325.27 billion~\cite{hoeffler2017costs}.
Furthermore, current countermeasures are often unsuccessful, with 
40\% of homicides worldwide going unsolved, and worsening clearance rates noted in countries such as the United States (US), Canada, Trinidad, and Tobago~\cite{sturup2015unsolved}.
As emphasized in the United Nation's 16th sustainable development goal (Peace and Justice), new approaches are urgently needed to improve crime prevention and investigation~\cite{un2023sustainable}.

Yet, crime often takes place indoors---behind closed doors and without witnesses---requiring careful, often painstaking  documentation to piece together what transpired.
In such cases, as noted in forensic literature, "the crime scene investigator must set one goal above all others: secure the integrity of the crime scene . . .  Lost or compromised evidence makes a crime scene investigator's job harder and can seriously damage investigations. Blood spatter patterns and fingerprints can be inadvertently smudged, footprints or tire treads walked on if care is not taken, and trace evidence can be scattered hither and yon by those unaware of its very presence".\footnote{\footurlContamination}
Furthermore, time is a critical factor: Evidence can degrade due to bacteria, heat, light, moisture, or mold~\cite{benson2003without}, and investigations are often delayed or cut short by understaffing or limited resources. As the saying goes, "when seconds count, the police are only minutes away."
Thus, to help preserve indoor crime scenes more carefully and automatically, we imagine that \emph{blimps} with cameras could be used to float over and record potentially sensitive evidence.

\begin{figure}
\centering
\includegraphics[width=.5\textwidth]{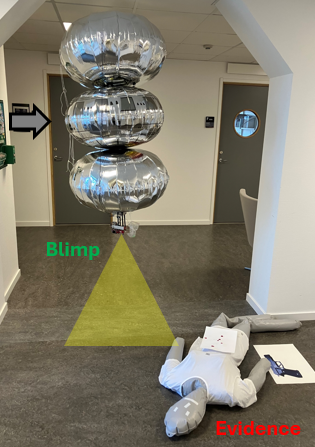}
\caption{Basic concept: a "floating camera" could be used to record sensitive evidence without destroying it.} \label{fig_results1}
\end{figure}

The term "blimp" traditionally refers to a non-rigid, lighter-than-air aircraft equipped with actuators and sensors within a “gondola” or “nacelle” to enable flight and monitoring.\footnote{\footurlDefinition}
Here, we adopt the term in a relaxed sense---e.g., allowing for some small rigid structure---in the belief that our ideas could be also broadly relevant for other kinds of aerostats, airships, dirigibles, and lighter-than-air unmanned aerial vehicles (LTA-UAVs), like hybrid dynastats or rotastats, zeppelins, and balloons.
While aerodynes such as drones could also aid in indoor crime scene analysis, blimps offer some unique advantages: safety, silence, stability (ability to hover over extended periods), sustainability (due to low cost, energy use, and emissions), soft landing, easy transport when deflated, and minimal training or licensing requirements for operators~\cite{rozhok2023modeling}.
Conversely, typical disadvantages---such as sensitivity to wind, low speed, and large size---could be less limiting in indoor environments, which are generally less windy and more compact, yet wide enough to be navigable by humans.

A challenge is that blimps intended to help solve difficult crimes in chaotic human environments would require a complex array of hardware and software capabilities.
To tackle this challenge, we adopted a rapid prototyping methodology that emphasizes "flexibility, possibility, and design insights as incubated subjectively through the designer", seeking to "cast our net widely"~\cite{zamfirescu2021fake}.
Thus, our contribution lies in reporting on some opportunities and limitations revealed during our design experience.

The remainder of the paper is organized as follows: In Section~\ref{section:related-work}, we summarize some previous work on blimps for security and indoor use.
Details from our exploration in Section~\ref{section:methods} are summarized in Section~\ref{section:discussion}, which also points out some next waypoints on our journey to supporting safe and just societies.

%\section{RELATED WORK}
\section{Related Work}
\label{section:related-work}

The basic potential of flying platforms for facilitating crime scene analysis has been explored, by manually piloting drones---or simulating their flight---to spot mock-up bloodstains, guns, knives, and bodies~\cite{urbanova2017using,georgiou2022uav,araujo2019multi}.
We have also previously presented an overview of tasks that a flying robot could perform at a crime scene~\cite{cooney2025nano}. 
However, Bucknell and Bassindale raised the alarm that downwash---wind from a drone's propellers---could disturb sensitive evidence~\cite{bucknell2017investigation}.
This could be problematic, since fibers and fine particles have played a crucial role in solving some important but challenging cases, such as those of the "Route 40" or "Green River" killers.\footnote{\footurlFiber}$^,$\footnote{\footurlParticles}
In their study, Bucknell and Bassindale concluded that drones should be flown at higher altitudes, based on experiments involving a Parrot AR.Drone 2.0 at varying heights~\cite{bucknell2017investigation}.
Nonetheless, wind could still be generated, especially in indoor settings with low ceilings or when examining elevated evidence on tables or shelves.
In contrast, a blimp that hovers or drifts slowly without using its propellers could generate less airflow---an assumption that motivates the present study.

More broadly, blimps have been previously considered for security-related applications:
For instance, Murphy and colleagues described how the police could use blimps to monitor and deter crime---an idea discussed in the city of Ogden in the US~\cite{murphy2013future}.\footnote{\footurlUtah}
Saiki et al. also proposed the use of a blimp to monitor disasters, describing control equations for a 12 m prototype with a stereo camera and LiDAR~\cite{saiki2011path}. 
As well, the Ukrainian Armed Forces are using tethered aerostats with infrared sensing from Aerobavovna and Kvertus to help with surveillance, communication, and drone deployment when an enemy drone is detected~\cite{trevithick2025balloon}.
The same article notes that China and the US have also proposed aerostats that can launch drone swarms---an idea that could also hold promise for analysis of larger crime scenes.

Various companies have also designed small indoor blimps, for entertainment:
For example, a 1994 patent (US5429542A) proposed a helium-filled, remote-controlled saucer toy; over the years, various toys have followed, such as the \emph{Megatech Blimp} in the early 2000s, a fish-shaped blimp called \emph{Air Swimmers} in 2011 and the \emph{NanoBlimp} in 2013.\footnote{\footurlAirSwimmers}$^,$\footnote{\footurlNanoBlimp}
As well, the company Festo has stirred interest with many imaginative creations: e.g.,   
\emph{Air\_ray}, a floating manta ray introduced in 2007;\footnote{\footurlAirRay}
\emph{AirJelly}, a jellyfish from 2008;\footnote{\footurlAirJelly}
\emph{AirPenguin}, a penguin from 2009;\footnote{\footurlAirPenguin}
\emph{SmartInversion}, a moving origami-style blimp from 2012;\footnote{\footurlSmartInversion}
\emph{eMotionSpheres}, a swarm of dancing spherical blimps with LEDs from 2014;\footnote{\footurlEmotionSpheres}
and \emph{FreeMotionHandling}, a spherical blimp that can pick up a bottle and hand it to a person, from 2016.\footnote{\footurlFreeMotionHandling}

Several research papers have also explored the design of novel indoor blimps for diverse applications:
For example, in 2012, we built a first flying humanoid robot prototype, \emph{Angel}, intended to safely interact with people~\cite{cooney2012designing}.\footnote{\footurlAngel} In addition to exploring how to plan safe motions and convey functional or emotional meanings through flight, we also experimented with two flight mechanisms, utilizing wings or propellers.
In 2015, St-Onge et al. made a uniquely-shaped cubic flying blimp, for indoor flight at art shows~\cite{st2015dynamic}. 
As well, in 2019, Yao et al. built a blimp to safely follow people indoors, detecting faces and two hand gestures via an RGB camera, and communicating via visual flight patterns and an LED display~\cite{yao2019autonomous}. 
That same year, Ferdous et al. built a blimp with low-current components, including an infrared (IR) sensor and inertial measurement unit (IMU), to enable usage over many hours~\cite{ferdous2019developing}.
In 2020, Seguin et al. used a convolutional neural network (CNN) to estimate the horizontal position and heading of a blimp from RGB video, along with a LiDAR to estimate height~\cite{seguin2020deep}.
Recently, in 2024, Pham et al. described a bio-inspired blimp with fins and an onboard IMU~\cite{pham2024controlling}.
Additionally, in 2024, Huang et al. designed an interesting 36" blimp with an OpenMV camera that can conduct remote measurement of important vital signs such as heart rate, respiration, or blood pressure via remote photoplethysmography; the blimp detects people with YOLO, approaches via PID control, and transmits data from stably detected foreheads~\cite{huang2024cost}.
Furthermore, in 2025, Xu et al. described a fish-shaped, flapping-wing blimp named \emph{Cuddle-Fish} that users enjoyed touching~\cite{xu2025cuddle}.
The same year, Hong and Tanaka presented unique walking humanoids with balloon torsos and articulated legs containing LEDs and various sensors, that can be controlled with gamepads or directed via external fans at an art installation~\cite{hong2025buoyant}.
While informative, none of these studies considered the unique challenges of crime scene analysis by a blimp.

With regard to functions for movement, documentation, and sensing to enable manual piloting, much prior work on blimps seems to have focused on control algorithms to deal with wind for particular embodiments~\cite{bhat2024review}---suggesting the usefulness of checking how easy it is to pilot a CSA blimp manually.
As well, videos captured by a blimp's camera require time to watch; given that various previous work has explored the 3D mapping of crime scenes~\cite{esposito2023advances}, it seemed useful if a video could be leveraged to generate a 3D model that can be immediately inspected.
Additionally, various other modalities could be used for documenting evidence, such as heat, sound, smell, or distances through solids (via radar or WiFi signals).
For example, thermal traces could reveal recent activity, like jettisoned or hidden weapons~\cite{munoz2025concealed} or drugs.

Looking further ahead, it could also be beneficial to explore the feasibility of automatically detecting and classifying evidence, and how a blimp itself could plan to move to acquire data. 
For example, we have previously used YOLO, to detect suspicious objects~\cite{cooney2024designing}.
Such a tool could also be used to gain insight into an important kind of evidence, blood.
Bloodstains are generally categorized into three main kinds: \emph{passive} (e.g., from drops falling under gravity), \emph{active} (e.g., from blood expelled from a victim's body or a weapon), or \emph{transfer} (e.g., from contact by a bloody hand or shoe).\footnote{\footurlBlood}
While Bergman et al. recently described using a CNN to discriminate between passive drip vs. active spatter bloodstains~\cite{bergman2022automatic}, we are not aware of work that has tackled classification across all three categories---possibly due to the sensitivity and scarcity of publicly available crime data.

In addition to interpreting what it "sees", an autonomous blimp should also plan how to move to acquire data.
The problem of how a robot should move to cover a given area is known as \emph{coverage path planning} (CPP). An NP-hard problem related to the Traveling Salesman Problem, CPP arises in a wide range of applications, from vacuum cleaning, to lawn mowing, snow removal, search-and-rescue, and aerial or underwater terrain exploration~\cite{jonnarth2023learning}.
For example, typical simplified algorithms used by robotic vacuum cleaners include snaking (also known as a zigzag or s-path algorithm---often combined with random walks), as well as spiraling and random motion~\cite{sorme2018comparison}. Among these, snaking was found to be efficient at achieving high coverage faster, with fewer unneeded passes---at the cost of increased turning and weakness to wheel or sensor errors that can occur in lower-cost robots. 
In more complex scenarios where optimized algorithms are desired, the state of the art increasingly involves deep reinforcement learning (RL; e.g., for vacuum cleaners~\cite{moon2022path}). 
Accordingly, various manual and autonomous capabilities appeared like they could be useful, inviting exploration.

%\section{METHODS}
\section{Methods}
\label{section:methods}

To start exploring potential capabilities, a basic platform was first required. With the aim of facilitating replication, cost-effective, standard-sized, off-the-shelf components and freely available software tools were used, as described below.

\subsection{Set-up}
\label{section:set-up}

\subsubsection{Hardware}
\label{section:hardware}

Various electronics components where used, such as those summarized in Table~\ref{tab:components}. 

\begin{table*}
  \caption{Electronics Components: Approx. Weight and Current}
  \label{tab:components}
  \begin{tabular}{ccl}
    \toprule
    Component&Weight&Current\\
    \midrule
    Raspberry Pi Zero 2 W  & 10 g & 460 mA\footnote{\footurlRPICurrent}$^,$\footnote{\footurlRPIWeight}\\
    ESP32-Cam  & 10 g & 180- 310 mA\footnote{\footurlESPCurrent}$^,$\footnote{\footurlESPWeight}\\
    TF-Luna & 5 g &  70 mA (x 3 = 210 mA)\footnote{\footurlLunaSpecs} \\
    MLX90640 & 3.5 g & 20 mA\footnote{\footurlMLXSpecs}\\
    Lipo Shim & 2.6 g & (15 $\mu$A quiescent)\footnote{\footurlLiPoShimSpecs}\\
  \bottomrule
\end{tabular}
\end{table*}

%as shown in Fig.~\ref{fig_blimp}
\emph{Computing.}
To obtain and wirelessly communicate sensor data, we set up a small Raspberry Pi Zero 2 W minicomputer (6.5 cm × 3.0 cm) with a 1 GHz quad-core 64-bit Arm Cortex-A53 CPU, 512 MB of SDRAM, and 2.4 GHz WiFi.\footnote{\footurlRPI}
As well, to control the blimp, an external laptop comprising an Intel i7 CPU @ 2.60GHz with 16 GB RAM and a 6GB GPU was used.

\emph{Sensors.}
%\subsubsection{Sensors}
\emph{RGB Camera}. A downward-facing ESP32-Cam was used to stream SVGA at 30fps from its 2 Megapixel OV2640 
camera.\footnote{\footurlESP}
%(FoV 66 degrees?) check
(Various alternatives exist, such as the OpenMV camera,\footnote{\footurlOpenMV} which although costing more, features more onboard processing capabilities.) 
 
%https://en.benewake.com/TFLuna/index.html
\emph{LiDARs}. To detect distances in front, to the side, and below the blimp (e.g., allowing wall-following or mapping), three TF-Lunas were used. TF-Luna is a small (3.5 cm x 2.1 cm x 1.3 cm, weight $\leq$ 5g) Time-of-Flight (TOF) single-point LiDAR based on an 850 nm Vertical Cavity Surface Emitting Laser (VCSEL), which detects distances within a range of 0.2 m to 8 m with 1 cm resolution and 6 cm accuracy.\footnote{\footurlLuna}
Again, alternatives existed, like the VL53l1X sensor with 4 m range---yet the TF-Luna's longer range felt like it could be useful in larger rooms or corridors.

\emph{Thermal camera}. Along the way, to explore possibilities for other modalities (e.g., detecting warm evidence that has recently been touched by a user), we attached an MLX90640 board. MLX90640 is a small (2.5 cm x 1.8 cm x 1.6 cm, 3.5 g) 768 pixel thermal camera with a 110$\degree$ x 70$\degree$ field of view, composed of a 24 x 32 array of IR thermal sensors that detect from -40$\degree$C to 300$\degree$C with an accuracy of $\pm$2$ \degree$C at a frame rate of 2 Hz.\footnote{\footurlThermal}
More powerful but costly alternatives such as FLIR exist.

\emph{Power.}
%\subsubsection{Power} 
To power the microcomputer and sensors, a "LiPo Shim" was used, which provides a maximum of 1.5 A continuous current from a 1 cell lithium polymer battery (a low battery warning occurs at 3.4 V, followed by automatic shutdown at 3.0 V).\footnote{\footurlLiPoShim}
With our two small batteries (700 mAh or 1000 mAh, approx. 15-20 g), operation was approximately 20 minutes.
As well, a separate 3 V CR2 battery was used to power the actuators. 

\emph{Actuation and Body.}
%\subsubsection{Actuation and Body}
To achieve motion without "reinventing the wheel", we modified an existing toy, comprising three DC motors attached to propellers and a radio receiver/transmitter.
The propellers allow left and right rotation, as well as upward, downward, forward, and backward translation.
To control the blimp via our own program, we connected the transmitter to our laptop via some common NPN bipolar transistors (2N2222) and an Arduino Uno.\footnote{\footurlArduino}
To lift the gondola, three balloons were used, each with approximately 80 cm diameter, 40 cm height, and a lift capacity of approximately 70-80 g when full (thus, 200-250 g in total). (Given the weight of helium and air (0.18 kg/m$^3$ vs. 1.29 kg/m$^3$), and a lift of 1.11 kg by a cubic meter of helium, this suggests there was somewhat less than a tenth of a cubic meter in each balloon.)\footnote{\footurlHelium}

During our design exploration, we observed that equipping small blimps with many sensors posed challenges:
\emph{Size}. 
Miniature toy units with regular latex balloons offered minimal payload capacity and lost helium quickly. Using two balloons with a larger toy improved lift, but the cumulative weight of sensors, boards, cables, tape, and power systems added up fast. High current demands from the microcomputer, sensors, and actuators required a large battery, and separate power supplies were adopted to avoid voltage spikes that could reset the microcontroller. 
To overcome these constraints and enable free exploration, our latest version became substantially larger, using three balloons (total height $\approx$ 150 cm with gondola and ballast).

\emph{Assembly}. 
Initial attachment methods using tape or velcro proved unreliable; gondolas detached or tore balloon surfaces. Unlike our earlier “Angel” design, where balloons were anchored to a table, we avoided heavy hooks this time by temporarily tucking balloons under tables to immobilize them. At one point, we used a 2.3 m corridor ceiling to assemble the blimp, but public access made it impractical---once, we glimpsed a passerby whacking our blimp playfully, causing the balloons to break apart. We later secured the blimp under a ceiling-mounted projector in our lab, tying balloon tails also to chairs as needed, and connecting balloons via capped carbon fiber rods held by modified wall hooks. Despite this, the top balloon sometimes detached, so we added set screws at both ends to secure the rods.

To prevent the gondola from detaching, we taped a plastic base to the bottom balloon, reinforced with threads to side hooks, and affixed the gondola using bolts and nuts. Sensors had to be mounted below the propellers, so we suspended a second lightweight cardboard gondola from the main one using rods. We also used ballast to fine-tune buoyancy, but noticed that battery changes (e.g., between 700 mAh and 1000 mAh) required recalibration. While testing propellers on a tabletop, we were also surprised when only briefly powering a motor caused the entire gondola to spin and hit a laptop, breaking a propeller---after which we made sure to tape down the gondola securely when testing.

\begin{figure}
\centering
\includegraphics[width=\textwidth]{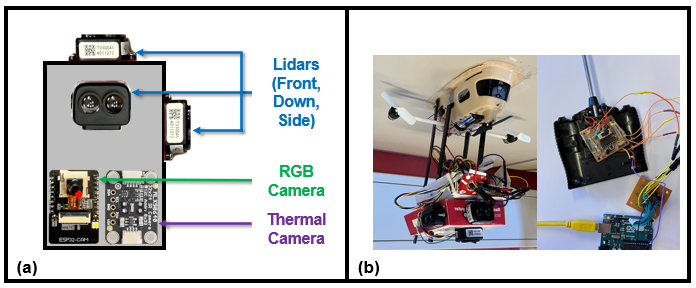}
\caption{Hardware: (a) Base of the lower gondola, showing sensor positions (b) gondola with 3 propellers, minicomputer, battery, and sensors on the left, and radio transmitter on the right} \label{fig_gondola}
\end{figure}

%\begin{figure}
%\centering
%\includegraphics[width=.5\textwidth]{images/transmitter.jpg}
%\caption{Transmitter} \label{fig_transmitter}
%\end{figure}

\subsubsection{Software}
\label{section:software}

\begin{figure}
\centering
\includegraphics[width=\textwidth]{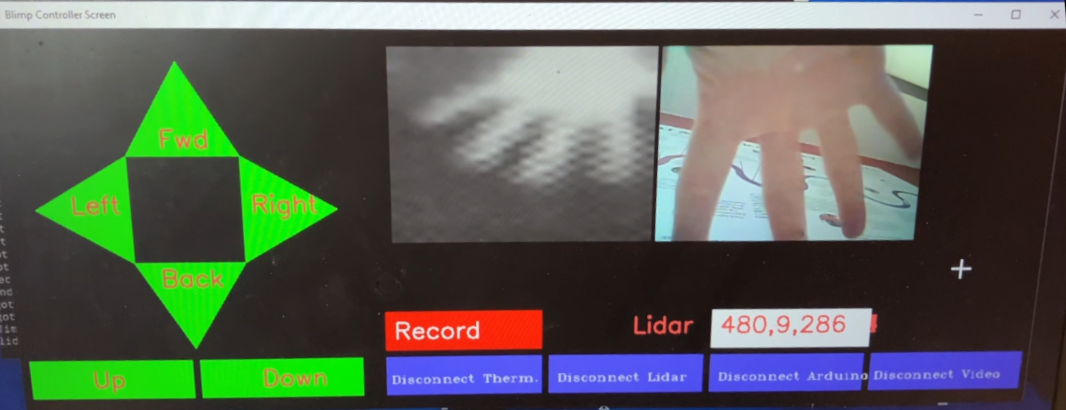}
\caption{User interface: Green buttons control the blimp's movements, blue buttons start threads to read sensors, and the red button enables recording. The text field shows LiDAR distances, and the videos show thermal (left) and RGB data (right).} \label{fig_ui}
\end{figure}

We also developed server and client programs to transmit LiDAR, RGB, and thermal data, as well as control signals between the gondola and our external laptop. 
TigerVNC was used to remotely develop the software on the Raspberry Pi.\footnote{\footurlTigerVNC} 
As well, we built a user interface with color-coded controls in Python and OpenCV,\footnote{\footurlOpenCV} shown in Fig.~\ref{fig_ui}.
Various Arduino functions were implemented to allow the blimp to move, but to reduce risks if communication was lost, we ultimately chose to use commands that initiate short bursts rather than continuous motion.

A key challenge to our exploration was a lack of information on how to set up reasonably realistic mock-up crime scenes: e.g., what kind of evidence is typical and where it is usually found. 
We found no publicly available datasets of crime scene layouts---likely due to privacy, legal, or safety concerns, or regional variances on how to categorize crimes. Aside from two scans of mock-up crime scene shared by Galanakis et al.~\cite{galanakis2021study} and the Swedish Police from their National Forensics Center (NFC) training facility, and a few mock-up sketches on the internet, most search results seemed to point to story-writing tools (e.g., “Murder Scene Generator”).\footnote{\footurlWriter} This was problematic, since, as the saying goes, "no two crime scenes are the same", blimps should be able to operate in different environments, and engineering evaluations typically require more than two or three trials (deep learning systems can even require thousands of training samples).

Thus, to support initial testing, we created two simplified simulation tools. The first lets users choose a crime type (e.g., homicide, assault, burglary, or arson) and floor plan (e.g., NFC Villa or our university’s 50 m$^2$ "HINT" smart home), then randomly generates evidence types and locations, using customizable probabilities. The second tool places all evidence in a single heap, filling a grid by size---larger items first.

\begin{figure}
\centering
\includegraphics[width=.45\textwidth]{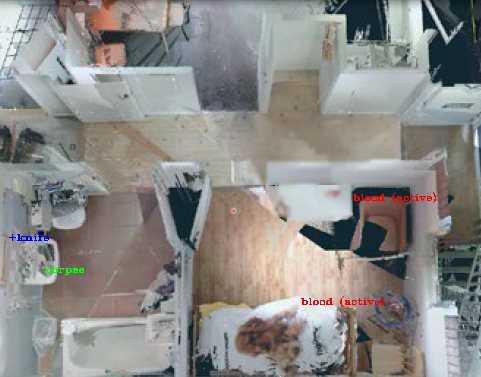}
\includegraphics[width=.45\textwidth]{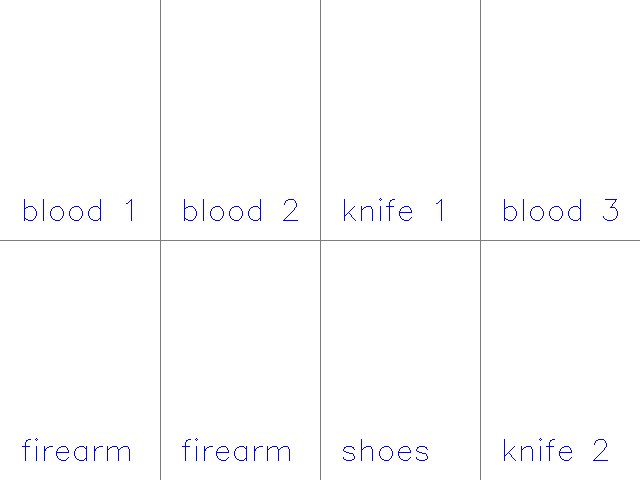}
\caption{Randomly generated crime scenes: (left) homicide at NFC Villa, (right) evidence heap} \label{fig_random_nfc}
\end{figure}

\subsection{Exploring Possibilities for Indoor CSA Blimps}
\label{section:sensing}

The developed platform was applied to check the questions identified in reviewing the literature: 
how much wind the blimp creates, how easy it is to control manually, how recordings of evidence can be further processed into 3D maps, and if thermal traces could be used.
Looking ahead, we also aimed to explore the feasibility of automatic detection, classification of bloodstains, and autonomous path planning by the blimp for data acquisition.
For evidence to use in testing, we gathered a photo of a firearm, two knives (one large and metallic, the other small and plastic), a pair of shoes, and three sheets with mock-up blood (red carmine dye---E120) to simulate passive, active, and transfer stains.

\subsubsection{Disturbance} To test the degree to which sensitive evidence might be disturbed by a blimp, we placed a Mastech MS6252A digital anemometer on the ground beside the mock-up evidence. As a result, the wind sensor consistently read 0.0 m/s, even when the blimp passed close ($\sim$20 cm) to the evidence.
For comparison, flying a typical small drone---the DJI Ryze Tello---over the evidence at 1.2 m resulted in wind speeds of 0.6 to 0.8 m/s, which notably caused most of our evidence to scatter (viz., the sheets of paper containing mock-up blood and the photo of a firearm).

\subsubsection{Sensing while Drifting} Next, we checked on the blimp's ability to move and sense evidence.
The blimp was manually controlled to obtain data ten times. In each trial, a random ordering of the evidence was calculated using our Python program, then the physical evidence was manually placed by a human.
As a result, on average, the blimp captured 5.2 of 7 objects (74.3\%) in its view.
Some objects were occasionally missed because the blimp was susceptible to drafts and small imbalances that sometimes caused it to veer.

\subsubsection{3D Mapping}
For easy viewing, we explored converting video from the blimp to a 3D model. To test the waters, a video from the blimp was processed using the free version of 3DF Zephyr to create a 3D textured mesh.\footnote{\footurlZephyr}
The result is shown in Fig.~\ref{fig_additional}(a).
Some areas appeared rough---for example, around the mannequin "corpse" on the left side of the image---since the free software version only processes the first 50 frames extracted from the video.
(Due to this limitation, we also sped up the video to ensure the entire scene was included.)
However, the short time required to generate a 3D map (approximately 3 minutes) seemed promising.

\subsubsection{Thermal Sensing} 
We also experimented with briefly touching a knife and the shoes before each trial controlling the blimp.
As a result, some warm areas were visible in the thermal data, as exemplified in Fig.~\ref{fig_additional}(b). However, in general, the heated objects---especially the smaller knife---seemed hard to distinguish due to the low resolution of our thermal sensor.
As well, a checkerboard pattern appeared due to the blimp's motion and the manner in which the MLX90640 sensor updates (in two passes, such that half of the pixels update slightly later). Thus, we believe that a higher-resolution sensor with a different update mechanism would be required to identify recently touched, non-metallic objects.
That said, the MLX90640 could still be useful for other purposes, such as detecting investigators (to follow) or hidden persons.

\subsubsection{Object Detection} 
Time and effort could be saved by automating key tasks such as detection of evidence.
Detection results could also be highlighted in a 3D map in reports to investigators, or used to focus the blimp's attention on areas requiring closer inspection.
As an initial check, we tried passing an image taken by the blimp to a pretrained generic YOLOv8l detection model\footnote{\footurlYOLO} and ChatGPT.\footnote{\footurlChatGPT}
As shown in Fig.~\ref{fig_additional}(c), the YOLO model correctly identified one knife, but a bloody handprint was mistaken as a horse, and other objects were not detected. These errors were likely due to low image resolution or dim lighting conditions (approximately 300 lux in our lab when the image was taken).
ChatGPT, by contrast, correctly detected all objects.
This seemed promising and suggested the usefulness of applying a generic foundation model, if hardware permits and a specific model fine-tuned on a sufficient number of images of firearms and blood is unavailable.

\subsubsection{Bloodstain Classification}  

\begin{figure}
\centering
\includegraphics[width=\textwidth]{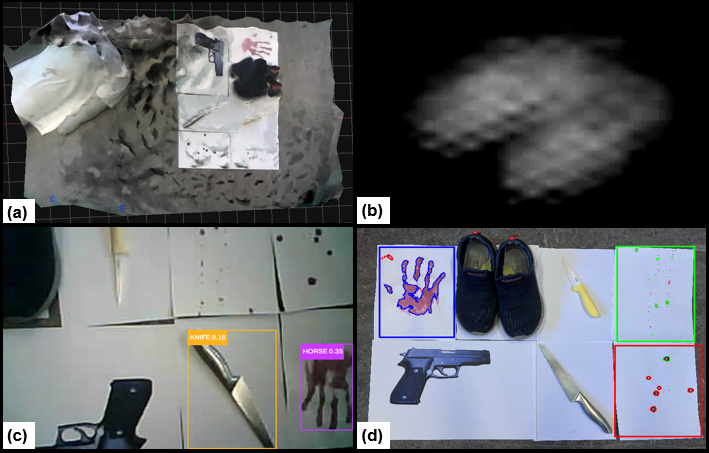}
\caption{Additional exploration: (a) photogrammetry/structure from motion,
(b) detecting warm shoes that a criminal might have recently used,
(c) detecting objects, 
(d) detecting blood pattern type (rectangles indicate ground truth, and contour colors indicate predicted class---red: passive dripping, green: active spatter, blue: transfer smears).} \label{fig_additional}
\end{figure}

To explore classifying the three main kinds of bloodstains, we tried processing a photo of our evidence "heap".
First, we used color-picking and contours to identify clusters of red pixels. 
Then, we assessed the area and eccentricity of an ellipse fitted to each detected contour of reasonable size.
We assumed that typical transfer stains (e.g., from a hand or foot) could be large compared to individual droplets, that passive drops should appear round (exhibit low eccentricity), and that active drops should appear elongated (have high eccentricity), as illustrated in Fig.~\ref{fig_additional}(d).
This resulted in a classification accuracy of 74.1\%, with 20 of 27 detected blood contours correctly classified.
Adding in false positives from color-picking (e.g., red tags on the shoes), the overall accuracy dropped to 69.0\% (20 of 29), which still seemed promising for initial prototyping.

\subsubsection{Coverage Path Planning for Crime Scenes} 
An automated blimp could operate without manual piloting.
Given that a full solution to the complex problem of autonomous coverage path planning for a CSA blimp would not be possible within the scope of the current paper, we constrained our brainstorming here to identifying key variables, requirements and potential strategies.
First, we propose that designs should consider the following factors:
\begin{itemize}
\item{\emph{Blimp's capabilities}
\begin{itemize}
\item{\emph{Movement precision}: If the blimp’s movements are imprecise, it should fly at a safer, farther distance to avoid collisions and plan more overlap to ensure coverage.}
\item{\emph{Sensor accuracy}: If sensors are less accurate, the blimp should fly closer to the evidence to capture sufficient detail.}
\end{itemize}
}
\item{\emph{Task}
\begin{itemize}
\item{\emph{Task urgency}: If time is constrained and the blimp is needed elsewhere, speed and efficiency can be prioritized. Else, in critical cases, more thorough scanning could be appropriate. For example, higher-detail documentation could require maneuvering beneath or around obstacles like tables and beds.}
\item{\emph{Task specification}: If the blimp is expected to also gather evidence, it should move closer to objects of interest.}
\end{itemize}
}
\end{itemize}

Thus, in general, we believe that indoor crime scene mapping by a blimp should meet the following requirements, listed in arbitrary order:
%In general, however, we believe that indoor blimp crime scene analysis is characterized by the following requirements:
\begin{itemize}
\item{\emph{Speed}: Timely mapping is desired, prior to contamination or degradation of evidence, also given battery life constraints (e.g., mapping a room should take an hour or less).}
\item{\emph{Overlap}: Some overlap (e.g., 25\%) is required for 3D reconstruction, but excessive motion could disturb the scene.}
\item{\emph{Navigability}: While some environments could be mostly empty and of simple rectangular shape, advanced capabilities could be required to navigate in complex domestic spaces with obstacles such as ceiling fans, lamps, chairs, or hallways, even if floor plans are known.}
\item{\emph{Efficiency}: The blimp should minimize turning and vertical movement to conserve battery life. Flying at a greater height than the tallest standing object could be efficient,  although indoor spaces constrain vertical movement more than outdoor ones.}
\item{\emph{Interaction}: Humans, animals, or dynamic obstacles such as other robots could unintentionally or intentionally obstruct the blimp. Safety is paramount.}
\item{\emph{Adaptability}: An advanced blimp could operate as part of a swarm, use an LLM to prioritize regions likely to contain evidence, and remain within a geofenced region to avoid disturbing people or objects.}
\end{itemize}

Given these requirements, both simple and advanced planning approaches seemed possible:
\begin{itemize}
\item{\emph{Basic approaches}: A snaking strategy combined with random walks, wall-following, or behavior-based robotics, at a collision-free height, could serve as a practical and intuitive baseline for early-stage testing.}
\item{\emph{Variable-height approaches}: 
For maximum detail, the blimp could calculate a close path just above each surface. Conversely, initial high-altitude scans could be followed by targeted low-altitude re-visits to interesting areas (e.g., where potential evidence has been detected).}
\item{\emph{AI-driven approaches}: Deep reinforcement learning (RL) could be used to mimic how investigators explore crime scenes. This could achieve high efficiency (minimizing flight time and avoiding redundant re-coverage), map quality (ensuring good feature overlap and loop closure, while avoiding poor viewing angles, low lighting, occlusions, shadows or reflective surfaces), room-agnosticism (functioning successfully in unknown or specially designed rooms or buildings, while avoiding hand-engineering edge cases), and robot-agnosticism (allowing different blimps to use the same process to learn how to move well). However, complex methods---such as deep RL---could also introduce challenges, including difficult implementation, large data requirements, extensive tuning requirements, and reduced explainability.
}
\end{itemize}

%\section{DISCUSSION}
\section{Discussion}
\label{section:discussion}

In this paper, we explored a new design idea, of using a blimp to document sensitive evidence at an indoor crime scene.
\begin{itemize}
\item{\emph{Proof of Concept (Hardware and Software)}. To our knowledge, our new prototype is the first small indoor blimp to carry a thermal camera, three LiDARs, or a minicomputer (rather than a microcontroller). In addition to discussing various hardware challenges, some software was also developed to pilot the blimp and "generate" crime scenes.}
\item{\emph{Manual operation}. As a result of our exploration, we found some further support that drones can disturb sensitive evidence and confirmed our premise that blimps can keep wind to a minimum.
As well, manually piloting was possible but imperfect, with an average of 74.3\% of objects captured in view in a single pass.
Furthermore, a 3D map was generated within a few minutes using free software, and some recommendations for thermal sensing were given.}
\item{\emph{Towards Automation}. All objects in the blimp's camera view were detected correctly by ChatGPT.
As well, the three main kinds of bloodstain were detected and classified with an accuracy of 74.1\% via color-picking and consideration of contour area and eccentricity.
Finally, we discussed path-planning considerations and strategies.
}
\end{itemize}

As well, a video and code have been made available.\footnote{\footurlYouTube}$^,$\footnote{\footurlGitHub}

\subsection{Limitations and Future Work}
\label{section:limitations}

The current study is limited by its exploratory design approach,  working with mock-up evidence in a lab: 
\begin{itemize}
\item{\emph{Unoptimized components and guesswork}. To reduce size and weight, surface mount or low power components (e.g., Raspberry Pi Pico instead of Zero 2 W), Dist-YOLO in place of LiDARs~\cite{vajgl2022distyolo}, or smaller batteries could be evaluated. 
Hybrid setups could also allow greater speed and payloads.
As well, our current reliance on a remote laptop---though common in related work---would pose security risks at a real crime scene, given that criminals could find a way to hack (intercept or jam) communications.
Thus, dedicated onboard hardware acceleration for computer vision is desired. For basic tasks, an NPU (neural processing unit) could suffice, such as the 1 GHz unit on OpenMV N6 that can run YOLO at 30 FPS.\footnote{\footurlOpenMVTwo}
Heavier Foundation Models or LLMs requiring GPUs (Graphic Processor Units) with sufficient VRAM (Video Random Access Memory) could also become feasible on future blimp platforms; e.g., 16 GB for a 30B LLM with 4-16 bit quantization.\footnote{\footurlLLM}
Our crime scene generation software is also heuristic-based and would benefit from training with real data to enable more realistic outputs.}
\item{\emph{Simplified approaches}. 
Control algorithms should be developed to help blimp operators. Given the low thrust of blimp propellers, we should clarify if drifting might not even be required in some contexts. Additional use cases for thermal sensing and alternative miniature thermal cameras should be explored, as well as other free 3D Mapping tools (that allow more than fifty photos). Operation in challenging conditions---such as darkness or smoke---should also be explored, for example using auxiliary lighting, low-light cameras, or near-infrared imaging. Ultimately, field evaluations with real crime scenes and data will be essential to develop real products and services.}
\item{\emph{Limited data and testing}. 
A more comprehensive study is required to identify the various kinds of evidence that could be important and how they can be detected.
Also, for classifying bloodstains, only a single image was used; a reasonably-sized dataset---ideally real---should be used to train and evaluate a more advanced machine learning model, to achieve better results. 
Furthermore, in-depth forensic interpretation of bloodstain composites could reveal valuable information such as weapon type, handedness, number of blows, sequence of injuries, sources of transfer stains (e.g., a hand or a shoe), or immediacy of death.
Path planning should also be implemented and tested; for this, an omnidirectional camera could also improve scene coverage while minimizing unnecessary flight, reducing the risk of disturbing sensitive evidence.}
\end{itemize}

Beyond documentation of evidence at indoor crime scenes, remote-controlled blimps could even aid in hostage or standoff situations by offering silent surveillance to inform response planning, and reduce investigator exposure to ambushes or danger.
Blimps could also benefit other sensitive applications where airflow or noise from conventional drones could be detrimental---for instance, munitions or chemical facilities with ignition risks, nuclear disaster sites where radioactive dust should not be stirred, avalanche or building sites in danger of collapsing, habitats of endangered animals sensitive to noise, or pharmaceutical cleanrooms or semiconductor plants where slight air movements could introduce flaws.
By pursuing such directions, our aim is to identify opportunities for technology to contribute to a more peaceful and happy society for all.

\begin{acknowledgments}
We gratefully acknowledge support from the Swedish Innovation Agency (Vinnova) for the project "AI-Powered Crime Scene Analysis" as well as various advice from the Swedish Police Authority (the initial idea for this paper stemmed from a conversation with Mikael Lilja). 
\end{acknowledgments}

%%
%% Define the bibliography file to be used
\bibliography{blimp_crime} %sample-sais

\appendix

\end{document}